# A Weak-Signal-Aware Framework for Subsurface Defect Detection: Mechanisms for Enhancing Low-SCR Hyperbolic Signatures

Wenbo Zhang[1], Zekun Long[*,2], Zican Liu[1], Yangchen Zeng[3], Keyi Hu[4]

[1]*Road and Traffic Engineering Research Center, Zhejiang Normal University, Jinhua, Vhina*

[2]*School of Information and Communication Technology, Griffith University, Brisbane, Australia*

[3]*School of Cyber Science and Engineering, southeast university, Nanjing, China*

[4]*School of Artificial Intelligence, Hainan Normal University, Haikou, China*

Email: {zhangwenbo, liuzican}@zjnu.edu.cn, zekun.long@griffithuni.edu.au, yangchenzeng@seu.edu.cn, huyike@hainnu.edu.cn

## ABSTRACT

Subsurface defect detection via Ground Penetrating Radar (GPR) is challenged by "weak signals"—faint diffraction hyperbolas with low signal-to-clutter ratios (SCR), high wavefield similarity, and geometric degradation. Existing lightweight detectors prioritize efficiency over sensitivity, failing to preserve low-frequency structures or decouple heterogeneous clutter. We propose WSA-Net, a framework designed to enhance faint signatures through physical-feature reconstruction. Moving beyond simple parameter reduction, WSA-Net integrates four mechanisms: Signal preservation using partial convolutions (PConv); Clutter suppression via heterogeneous grouping attention (LWGA); Geometric reconstruction (SCConv) to sharpen hyperbolic arcs; Context anchoring (CAA) to resolve semantic ambiguities. Evaluations on the RTSTdataset show WSA-Net achieves 0.6958 mAP@0.5 and 164 FPS with only 2.412 M parameters. Results prove that signal-centric awareness in lightweight architectures effectively reduces false negatives in infrastructure inspection.

## 1 Introduction

Subsurface defect detection via Ground Penetrating Radar (GPR) is essential for urban infrastructure, identifying voids, moisture, and delamination [1, 2]. By capturing reflections from dielectric discontinuities, GPR provides a non-invasive view of roadbeds and tunnels [3]. However, complex urban environments challenge inspection reliability due to "weak signals." We define these by three intrinsic challenges: (i) Low SCR, where faint reflections are submerged in heterogeneous noise [4]; (ii) High Wavefield Similarity, where different defects produce near-identical patterns [5]; and (iii) Geometric Atrophy, where hyperbolic wavefronts appear fragmented due to attenuation [6].

Deep learning, especially CNNs and YOLO, has automated GPR interpretation [7, 8]. However, lightweight frameworks optimized for optical imagery lack explicit modeling for GPR wavefields. Model compression for portable systems often discards subtle feature maps, exacerbating signal loss. Existing attention or fusion methods treat GPR data as generic textures rather than physical anomalies, failing to decouple non-uniform clutter [9-12].

To bridge this gap, we propose WSA-Net, a framework designed to enhance faint signatures through physical-feature reconstruction. Our contributions are:

1.Signal Preservation Mechanism: We adopt a Fasternet_t0 backbone utilizing Partial Convolution (PConv). Unlike aggressive downsampling in standard detectors, this mechanism minimizes computational redundancy while maintaining the integrity of low-frequency wavefield structures, ensuring that faint hyperbolic peaks are not filtered out during feature extraction.

2.Heterogeneous Clutter Suppression: We introduce Light-Weight Grouped Attention (LWGA). By employing heterogeneous feature grouping and sparse sampling, this mechanism explicitly decouples non-uniform background noise from micro-target reflections, significantly enhancing the model's sensitivity to low-SCR signatures.

3.Geometric Reconstruction and Contextual Anchoring: We integrate Spatial-Channel Reconstruction Convolution (SCConv) to sharpen geometrically degraded edges of subsurface defects, combined with Context Anchor Attention (CAA) to capture long-range spatial dependencies. Together, these mechanisms resolve semantic ambiguities and restore the connectivity of fragmented hyperbolic arcs.

## 2 Methodology

### 2.1 Weak-Signal-Aware Design Principles

GPR weak-signal detection is constrained by electromagnetic wavefields under strong attenuation and heterogeneous conditions. Unlike optical targets with stable textures, subsurface defects manifest as faint diffraction responses requiring signal energy preservation, clutter separation, and geometric coherence.

Existing lightweight detectors suffer from three deficiencies: Energy Loss: Aggressive compression suppresses low-frequency components, causing faint reflections to vanish during early feature extraction; Clutter Interference: Non-stationary subsurface clutter (from moisture or dielectric variations) overlaps spectrally with targets, making decoupling via uniform convolution ineffective; Geometric Fragmentation: Conventional detectors lack mechanisms to reconstruct global hyperbolic curvature, which often becomes blurred or fragmented under attenuation.

Accordingly, we formulate weak-signal-aware detection as a unified modeling problem composed of three interdependent objectives: (i) preservation of faint wavefield energy, (ii) statistical decoupling of heterogeneous noise, and (iii) restoration of degraded geometric consistency. Rather than constructing the detector as a simple aggregation of architectural components, the proposed framework establishes explicit correspondences between these modeling objectives and dedicated signal-centric mechanisms.

### 2.2 Overall Architecture

Based on the above principles, we propose WSA-Net, a weak-signal-aware detection framework instantiated on a lightweight one-stage detection architecture [12] for real-time subsurface inspection. As illustrated in Figure 1, the framework consists of three main stages: a signal-preservation backbone, a weak-signal-aware feature fusion neck, and multi-scale decoupled detection heads.

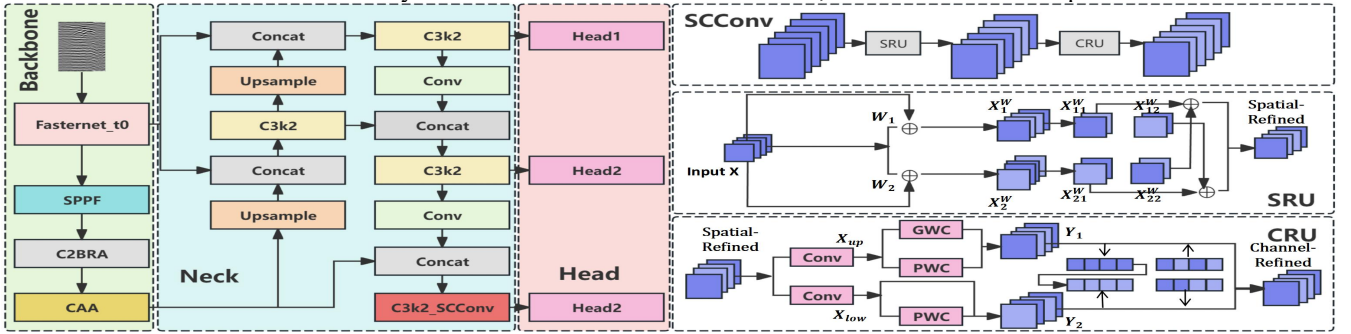

**Figure 1:** Architecture of WSA-Net. The framework integrates a Fasternet_t0 backbone, CAA, and an SCConv-based neck. The right panel details the SCConv module, comprising Spatial (SRU) and Channel (CRU) Reconstruction Units.

### 2.3 Signal Preservation Mechanism

Preservation of faint wavefield energy is the first prerequisite for weak-signal detection. In GPR B-scans, weak subsurface reflections are predominantly encoded in low-frequency structural components rather than high-frequency textures. However, conventional convolutional backbones often suffer from channel redundancy and excessive downsampling, leading to irreversible attenuation of weak responses.

To address this issue, WSA-Net employs a Fasternet_t0 [8] backbone (Figure 2) based on Partial Convolution (PConv). Unlike standard convolution, PConv applies spatial convolution only to a subset of channels while propagating the remaining channels through identity mapping. Formally, given an input feature map $X \in \mathbb{R}^{H \times W \times C}$, PConv operates as:

$$Y = \text{Concat}(\text{Conv}(X_{1:c}), X_{c+1:C}) \quad (1)$$

where convolution is applied to only $c$ channels and the remaining $C - c$ channels are preserved.

This design significantly reduces memory access cost and computational redundancy while maintaining high-throughput transmission of low-frequency wavefield information. Compared with full convolution, the floating-point operations are reduced by approximately 1/16 under typical configurations, enabling millisecond-level inference without sacrificing signal integrity. By limiting early-stage signal loss, the backbone ensures that faint hyperbolic energy is retained for subsequent modeling stages.

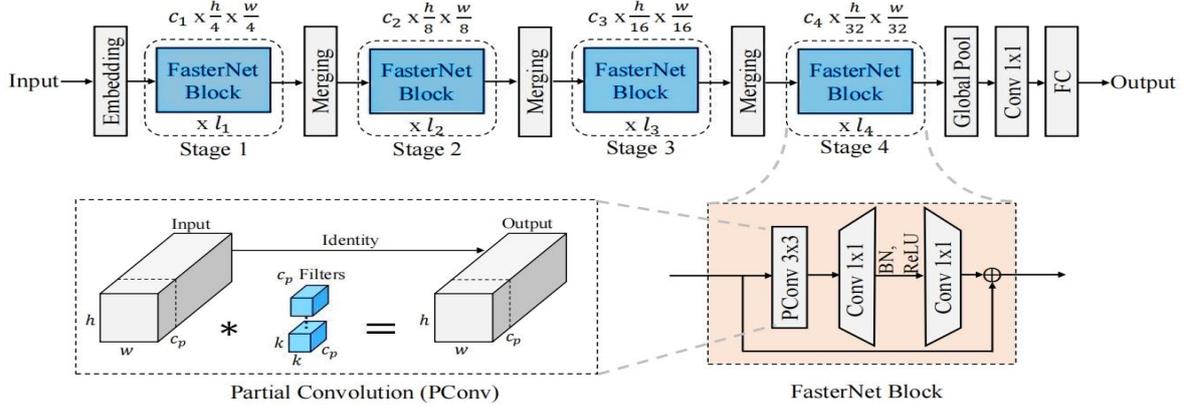

**Figure 2:** Structure of Fasternet_t0. The architecture comprises four hierarchical stages. The detailed inset illustrates the FasterNet Block, which employs PConv followed by PWConv for efficient feature extraction.

## 2.4 Heterogeneous Noise Suppression

In practical GPR environments, weak defect reflections are frequently masked by high-amplitude background clutter whose statistical properties vary spatially and spectrally. Conventional attention mechanisms compute dense global dependencies over all spatial positions, which often results in feature collapse: dominant clutter responses monopolize attention weights and suppress low-energy targets.

To achieve effective signal–clutter decoupling, we introduce Light-Weight Grouped Attention (LWGA), which explicitly models heterogeneity in feature statistics. Instead of treating the input feature map as a homogeneous entity, LWGA [9] partitions it into multiple subspaces, each responsible for suppressing or enhancing different interference patterns (Figure 3).

Specifically, the input tensor is decomposed into four complementary groups: Gate Point Attention (GPA): isolates sparse point-like anomalies and filters stochastic background noise; Regular Local Attention (RLA): enhances local texture continuity critical for weakened diffraction traces; Sparse Medium-range Attention (SMA): captures intermediate dependencies using sparse sampling; Sparse Global Attention (SGA): models long-range correlations through Top-K global feature interaction.

The outputs of all groups are adaptively fused as

$$F_{LWGA} = \sum_{i=1}^{} \alpha_i F_i \quad (2)$$

where $F_i$ denotes the output of each attention branch and $\alpha_i$ represents learned fusion weights.

By operating on heterogeneous subspaces and employing sparse sampling strategies, LWGA reduces computational complexity from quadratic to linear with respect to sampled tokens. More importantly, it prevents dominant clutter energy from overwhelming weak reflections, thereby significantly enhancing sensitivity under low-SCR conditions.

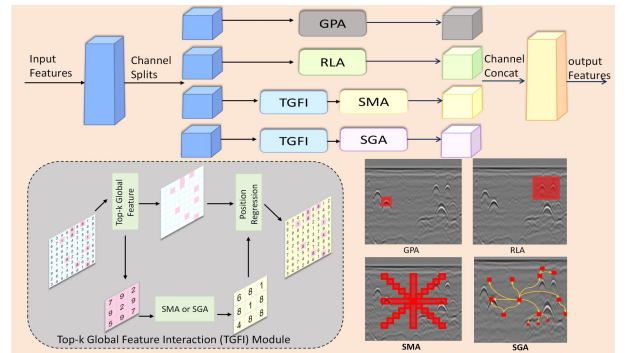

**Figure 3:** Illustration of the LWGA module. GPA, RLA, SMA and SGA denote gate point attention, local attention, sparse medium-range attention and sparse global attention.

## 2.5 Geometric Consistency Restoration

Subsurface defects in GPR data are primarily characterized by diffraction-induced hyperbolic wavefronts whose geometric continuity encodes critical physical information. Under strong attenuation and dispersion, these wavefronts often appear fragmented or blurred, causing conventional detectors to misinterpret them as isolated noise.

To restore degraded geometric structures, WSA-Net integrates Spatial–Channel Reconstruction Convolution (SCConv) within the feature fusion neck. SCConv [10] (Figure 1) consists of two complementary units. The Spatial Reconstruction Unit (SRU) utilizes Group Normalization responses to distinguish informative spatial patterns from redundancy and performs cross-reconstruction to enhance edge gradients. The Channel Reconstruction Unit (CRU) adopts a split–transform–merge strategy combining group-wise and point-wise convolutions, enabling effective reorganization of channel dependencies.

While SCConv sharpens local structural details, global curvature continuity requires contextual anchoring beyond local receptive fields. Therefore, a Context Anchor Attention (CAA) module is incorporated at the terminal stage of the backbone. CAA [11] (Figure 4) employs a pooling convolution activation paradigm with strip depth-wise convolutions to approximate global attention at minimal computational cost. By recalibrating feature responses according to long-range spatial context, CAA reinforces the integrity of hyperbolic wavefronts across fragmented regions.

Together, SCConv and CAA establish a dual-level geometric modeling mechanism: local reconstruction enhances edge clarity, while global anchoring enforces curvature consistency. This synergy enables the detector to perceive degraded diffraction signatures as coherent physical structures rather than disconnected local patterns.

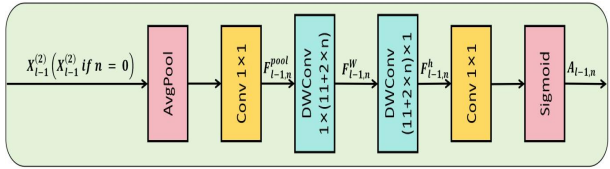

**Figure 4**：Schematic of the CAA module. The architecture follows a pooling-convolution-activation paradigm, utilizing strip depth-wise convolutions to capture long-range dependencies and a Sigmoid function for adaptive feature recalibration.

## 2.6 Mechanism-Driven Integration

Through the above design, WSA-Net establishes an explicit correspondence between weak-signal modeling objectives and architectural mechanisms:

Signal preservation is ensured by the partial-convolution-based backbone that minimizes early-stage energy loss;

Noise suppression is achieved through heterogeneous grouped attention that decouples clutter from weak targets;

Geometric consistency restoration is realized via spatial–channel reconstruction and global contextual anchoring.

By embedding these mechanisms within a lightweight detection architecture, the proposed framework aligns network behavior with the intrinsic physical properties of GPR wavefields. As a result, WSA-Net achieves robust weak-signal perception under severe clutter while maintaining real-time efficiency suitable for large-scale urban subsurface inspection.

## 3. Dataset Construction

### 3.1 Quantitative Definition of Weak Signals

We define "weak signals" in the RTSTdataset using four semi-quantitative criteria: Low Contrast ($C<0.2$): The peak amplitude of the hyperbolic reflection relative to the local background clutter, defined as $C = (A_{sig} - A_{bg})/(A_{sig} + A_{bg})$. Deep Burial & High Attenuation: Targets located at depths depths>1.5 m with energy dissipation > 40 dB. Clutter Dominance (SCR<-5 dB), where noise energy exceeds target reflection energy. Geometric Atrophy: Continuity rate< 60%, where the wavefront is fragmented into non-rigid segments due to dielectric interference.

### 3.2 Hybrid Dataset: Real-Virtual Complementary Strategy

The RTSTdataset (4,587 images) is constructed using a synergistic approach. While 3,587 high-resolution profiles provide real-world complexity, the 1,000 gprMax simulation samples are integrated not merely for data augmentation, but as explicit physical priors. These simulated wavefields offer: 1. Noise-Free Geometric Reference: Clean wavefronts for SCConv to learn edge reconstruction. 2. Controlled Parameter Variance: Moisture and depth variations to bound LWGA clutter decoupling. 3. Pixel-Perfect Ground Truth: Ensuring that boundary reconstruction loss is guided by exact physical dimensions, rectifying the boundary drift common in real-world manual labeling.

Defect statistics are summarized in Table 1. Labeling utilized a semi-automated workflow via X-AnyLabeling [13] with expert refinement.

**Table 1.** Defect distribution in the RTSTdataset

| Defect Type | Count |
|---|---|
| Cavity | 4852 |
| Void | 4007 |
| Loose | 2437 |
| Water-rich | 1209 |

As illustrated in the yellow bounding boxes in Figure 5, subsurface signatures often manifest as fragmented, low-contrast diffraction hyperbolas. Due to high dielectric similarity, defects exhibit subtle morphological differences and highly similar wavefield patterns. These micro-features are easily submerged by soil attenuation and heterogeneous clutter, leading to significant inter-class similarity and intra-class variation.

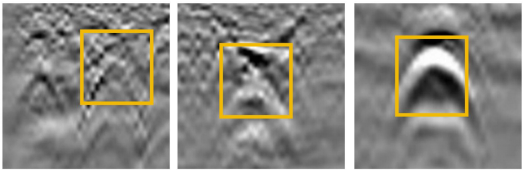

**Figure 5** Visualization of weak signal characteristics in GPR B-scans. Representative weak signatures showing fragmented hyperbolic arcs and high inter-class similarity under intense clutter

### 3.3 Generalization Benchmark: Roboflow-GPR Dataset
To evaluate cross-domain robustness, we utilize the public Roboflow-GPR (KSLMA) Dataset (976 images). This uncontrolled benchmark features significant deviations in equipment and clutter compared to RTSTdataset, serving to validate whether the model captures intrinsic hyperbolic wavefield characteristics rather than overfitting to specific textures.

### 3.4 Data Preprocessing
The dataset is partitioned into training, validation, and testing sets (7:2:1). To mitigate geometric distortions and simulate reflection variations under diverse roadbed conditions, we applied geometric transformations (translation, rotation, scaling) and photometric perturbations (brightness, contrast). This enhances model robustness against complex subsurface noise.

## 4 Results and Discussion
### 4.1 Comparative Experiments
Benchmark comparisons against ten architectures (Table 3) identify WSA-Net as the top performer (0.6958 mAP@0.5). Crucially, while heavy models like DINO [14] and Faster R-CNN [15] significantly underperform (~0.55 mAP), WSA-Net's superior Recall and Precision balance validates that our lightweight strategy avoids overfitting to the high-amplitude background noise common in low-SNR GPR data. As shown in Fig. 6, WSA-Net resolves the boundary drift and false negatives observed in baselines. This demonstrates physical consistency: by leveraging SCConv's edge reconstruction and CAA's global context, the framework accurately captures the non-rigid hyperbolic deformations of weak signals that are typically discarded by generic detectors.

**Table 3.** Comparative performance of WSA-Net and state-of-the-art models

| Model | P | R | Param /M | FLOPs /$10^9$ | FPS | mAP@0.5 |
|---|---|---|---|---|---|---|
| YOLOv8n[16] | 0.6823 | 0.6512 | 2.686 | 6.8 | 143 | 0.6639 |
| YOLOv8s[16] | 0.7011 | 0.6703 | 9.830 | 23.4 | 96 | 0.6821 |
| YOLOv11n[12] | 0.6909 | 0.6598 | 2.583 | 6.3 | 152 | 0.6719 |
| YOLOv11s[12] | 0.7015 | 0.6704 | 9.415 | 21.3 | 98 | 0.6821 |
| YOLOv12n[17] | 0.6745 | 0.6435 | 2.540 | 6.3 | 150 | 0.6552 |
| YOLOv12s[17] | 0.7009 | 0.6698 | 9.196 | 21.2 | 99 | 0.6817 |
| DINO[14] | 0.5698 | 0.5394 | 47.550 | 256 | 22 | 0.5500 |
| TOOD[18] | 0.5884 | 0.5576 | 32.030 | 180 | 26 | 0.5690 |
| RetinaNet[9] | 0.5532 | 0.5235 | 36.430 | 190 | 25 | 0.5340 |
| Cascade[20] | 0.5938 | 0.5624 | 69.229 | 219 | 18 | 0.5740 |
| Faster-RCNN[15] | 0.5726 | 0.5412 | 41.374 | 192 | 24 | 0.5530 |
| **WSA-Net(Ours)** | **0.7385** | **0.6692** | **2.412** | **4.2** | **164** | **0.6958** |

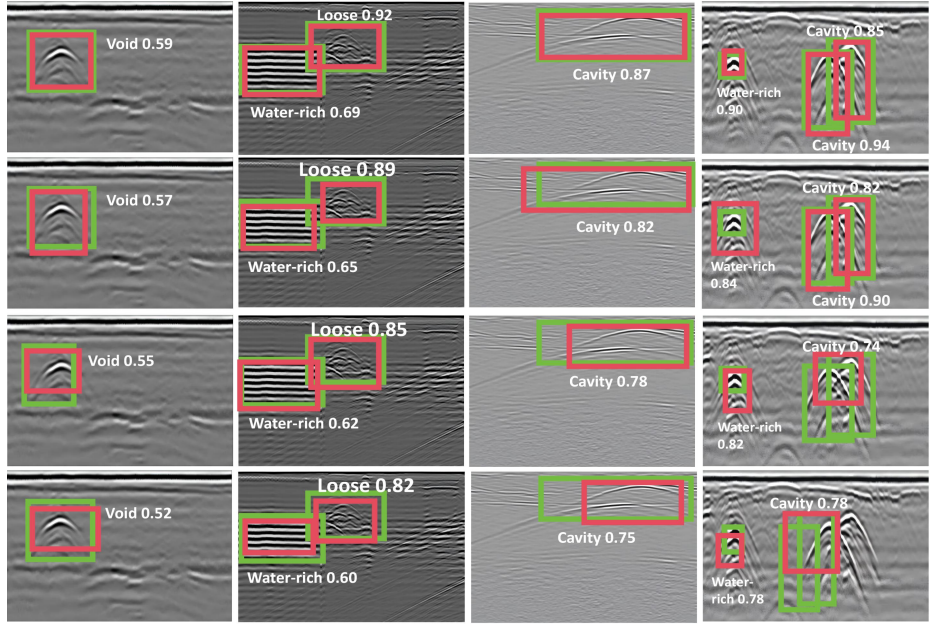

**Figure 6:** Visualization comparison between WSA-Net and YOLO series detectorVisualization and presentation of results.

## 4.2 Ablation Study

A stepwise ablation (Table 4) deconstructs how each mechanism contributes to weak-signal awareness: Signal Preservation: Replacing the backbone with Fasternet_t0 confirms that PConv eliminates computational redundancy while maintaining the signal throughput necessary for low-frequency features. Noise Suppression: The introduction of LWGA triggered a significant 5.4% Precision jump. This verifies that the heterogeneous grouping strategy effectively filters out non-uniform background clutter, preventing the feature collapse typically caused by uniform attention mechanisms. Geometric Consistency: The integration of SCConv and CAA primarily drove Recall improvements. By sharpening blurred edges and providing global context, these modules ensure that degraded hyperbolic arcs are perceived as coherent targets rather than isolated noise.

Heatmap analysis (Figure 7) further demonstrates physical consistency: unlike the baseline which reacts to irrelevant textual noise, WSA-Net's activation precisely targets the structural boundaries of subsurface defects.

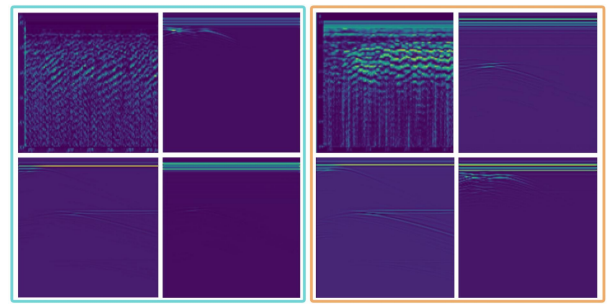

**Figure 7.** Heatmap comparison between YOLOv11 and WSA-Net.

**Table 4.** Ablation results of WSA-Net components

| base | Fasternet_t0 | SCConv | CAA | LWGA | $P$ | $R$ | Param | FLOPs | mAP |
|---|---|---|---|---|---|---|---|---|---|
| ✓ | | | | | 0.6909 | 0.6598 | 2.583 | 6.3 | 0.6719 |
| ✓ | ✓ | | | | | 0.6955 | 0.6612 | 2.228 | 5.4 | 0.6829 |
| ✓ | ✓ | ✓ | | | | 0.6942 | 0.6645 | 2.398 | 6.2 | 0.6788 |
| ✓ | ✓ | | ✓ | | | 0.6980 | 0.6625 | 2.652 | 6.4 | 0.6710 |
| ✓ | ✓ | | | ✓ | | 0.7280 | 0.6610 | 1.837 | 5.3 | 0.6827 |
| ✓ | ✓ | ✓ | | ✓ | | 0.7320 | 0.6640 | 2.056 | 4.7 | 0.6890 |
| ✓ | ✓ | ✓ | ✓ | ✓ | | **0.7385** | **0.6692** | **2.412** | **4.2** | 0.6958 |

### 4.3 Cross-Domain Generalization Assessment

To verify whether WSA-Net masters intrinsic physical representations rather than dataset-specific textures, we conducted zero-shot testing on the Roboflow-GPR (KSLMA) benchmark. This "uncontrolled" domain features significant statistical deviations in dielectric constants relative to our training set.

Without fine-tuning, WSA-Net achieved a 3.67% improvement in mAP@0.5 over the baseline (Table 5). Crucially, the LWGA module maintained a Precision of 0.6125 by suppressing heterogeneous clutter, effectively minimizing false alarms in unfamiliar geological contexts. Simultaneously, the model maintained a Recall of 0.5478, demonstrating its ability to reconstruct the underlying hyperbolic geometry across different equipment frequencies. These results underscore WSA-Net's readiness for large-scale engineering deployment where reliable weak-signal capture is paramount.

**Table 5.** Generalization performance on a public benchmark

| Model | Precision(P) | Recall(R) | mAP@0.5 | Params(M) | FLOPs(G) | FPS |
|---|---|---|---|---|---|---|
| YOLOv11n[14] | 0.5642 | 0.5103 | 0.5215 | 2.583 | 6.3 | 152 |
| WSA-Net | **0.6125** | **0.5478** | **0.5582** | **2.412** | **4.2** | **164** |

### 5 Conclusion

Our findings challenge the prevailing reliance on generic deep learning architectures for subsurface sensing, demonstrating that the robust detection of faint signatures requires a fundamental paradigm shift from visual texture recognition to physics-aware wavefield modeling. In this work, we proposed WSA-Net, a framework that addresses the intrinsic "weak signal" bottleneck—characterized by low SCR, clutter entanglement, and geometric atrophy—not through simple model scaling, but by embedding signal-centric mechanisms directly into the detection pipeline.

By integrating heterogeneous clutter decoupling (LWGA) and geometric wavefront reconstruction (SCConv), we validated that explicitly counteracting physical signal degradation is far more effective than generic feature extraction for resolving the "feature collapse" in low-SNR environments. Crucially, the weak-signal-aware methodology established here transcends the specific application of GPR. It offers a generalized, scalable blueprint for other sensing domains governed by similar wave propagation physics, such as Synthetic Aperture Radar (SAR), underwater sonar, and seismic exploration, where target visibility is strictly compromised by attenuation and scattering. Future integration of these mechanisms with temporal sequence analysis and edge-computing hardware promises to redefine the precision limits of real-time structural health monitoring in complex urban infrastructures.